\pdfoutput=1

\documentclass[11pt]{article}

\usepackage{EMNLP2023}

\usepackage{times}
\usepackage{latexsym}
\usepackage{booktabs}
\usepackage{arydshln}

\usepackage[T1]{fontenc}

\usepackage[utf8]{inputenc}

\usepackage{microtype}

\usepackage{inconsolata}

\usepackage{amsmath}
\usepackage{multirow}
\usepackage{graphicx}
\usepackage{subfigure}
\usepackage{algorithm}
\usepackage{algpseudocode}
\usepackage{enumitem}
\usepackage{tcolorbox}

\usepackage{colortbl}
\definecolor{Ocean}{RGB}{129,194,234}

\usepackage{color, soul}
\definecolor{tri_red}{RGB}{187,39,26}
\definecolor{tri_blue}{RGB}{75,119,209}
\definecolor{tri_green}{RGB}{120,166,90}
\definecolor{pipeline_red}{RGB}{187,39,26}
\definecolor{pipeline_green}{RGB}{71,116,44}
\definecolor{table_ocean}{RGB}{229,242,250}
\sethlcolor{table_ocean}

\usepackage{cleveref}
\crefformat{section}{\S#2#1#3}
\crefformat{subsection}{\S#2#1#3}
\crefformat{subsubsection}{\S#2#1#3}
\crefrangeformat{section}{\S#3#1#4 to~\S#5#2#6}
\crefmultiformat{section}{\S#2#1#3}{ and~\S#2#1#3}{, #2#1#3}{ and~#2#1#3}
\Crefformat{figure}{#2Fig.~#1#3}
\Crefmultiformat{figure}{Figs.~#2#1#3}{ and~#2#1#3}{, #2#1#3}{ and~#2#1#3}
\Crefformat{table}{#2Tab.~#1#3}
\Crefmultiformat{table}{Tabs.~#2#1#3}{ and~#2#1#3}{, #2#1#3}{ and~#2#1#3}
\Crefformat{appendix}{#2Appx.~\S#1#3}
\crefformat{algorithm}{Alg.~#2#1#3}
\Crefformat{equation}{#2Eq.~#1#3}

\newcommand{\stitle}[1]{\vspace{1ex} \noindent{\bf #1.}}

\usepackage{xspace}
\newcommand{\MODEL}{\mbox{\textsc{MetaScale}}\xspace}

\usepackage{graphicx}
\usepackage{caption}
\usepackage{subcaption}
\definecolor{newpink}{HTML}{F4CCCC}
\definecolor{newgreen}{HTML}{D9EAD3}
\definecolor{newyellow}{HTML}{FFF2CC}
\definecolor{newblue}{HTML}{E5F6FC}

\newcommand{\reducedstrut}{\vrule width 0pt height .9\ht\strutbox depth .9\dp\strutbox\relax}
\newcommand{\pink}[1]{%
  \begingroup
  \setlength{\fboxsep}{0pt}%
  \colorbox{newpink}{\reducedstrut#1\/}%
  \endgroup
}
\newcommand{\green}[1]{%
  \begingroup
  \setlength{\fboxsep}{0pt}%
  \colorbox{newgreen}{\reducedstrut#1\/}%
  \endgroup
}
\newcommand{\blue}[1]{%
\begingroup
  \setlength{\fboxsep}{0pt}%
  \colorbox{newblue}{\reducedstrut#1\/}%
  \endgroup
}

\title{MetaScale: Test-Time Scaling with Evolving Meta-Thoughts}

\author{
Qin Liu\textsuperscript{\rm 1}~~~
Wenxuan Zhou\textsuperscript{\rm 2}~~~
Nan Xu\textsuperscript{\rm 2}~~~
James Y. Huang\textsuperscript{\rm 2}~~~
Fei Wang\textsuperscript{\rm 2}\\
\textbf{Sheng Zhang}\textsuperscript{\rm 3}~~~
\textbf{Hoifung Poon}\textsuperscript{\rm 3}~~~
\textbf{Muhao Chen}\textsuperscript{\rm 1}\\
{\textsuperscript{\rm 1}}University of California, Davis~~~
{\textsuperscript{\rm 2}}University of Southern California~~~
{\textsuperscript{\rm 3}}Microsoft Research\\
\texttt{\{qinli, muhchen\}@ucdavis.edu}~~~
\texttt{\{zhouwenx, nanx, huangjam, fwang598\}@usc.edu}\\
\texttt{\{shezhan, hoifung\}@microsoft.com}\\
  }

\begin{document}
\maketitle

\begin{abstract}
One critical challenge for large language models (LLMs) for making complex reasoning is their reliance on matching reasoning patterns from training data, instead of proactively selecting the most appropriate cognitive strategy to solve a given task. 
Existing approaches impose fixed cognitive structures that enhance performance in specific tasks but lack adaptability across diverse scenarios. 
To address this limitation, we introduce \MODEL, a test-time scaling framework based on \emph{meta-thoughts}---adaptive thinking strategies tailored to each task. \MODEL initializes a pool of candidate meta-thoughts, then iteratively selects and evaluates them using a multi-armed bandit algorithm with upper confidence bound selection, guided by a reward model. 
To further enhance adaptability, a genetic algorithm evolves high-reward meta-thoughts, refining and extending the strategy pool over time. By dynamically proposing and optimizing meta-thoughts at inference time, \MODEL improves both accuracy and generalization across a wide range of tasks. Experimental results demonstrate that \MODEL consistently outperforms standard inference approaches, achieving an $11\%$ performance gain in win rate on Arena-Hard for GPT-4o, surpassing o1-mini by $0.9\%$ under style control.
Notably, \MODEL scales more effectively with increasing sampling budgets and produces more structured, expert-level responses.

\end{abstract}


    

\section{Introduction}
Large language models (LLMs; \citealt{achiam2023gpt,dubey2024llama,guo2025deepseek}) have demonstrated remarkable capabilities across a wide range of cognitive tasks.
With minimal or no guidance, they can perform human-like multi-step thinking processes~\cite{kojima2022large,wang2024chain}.
However, one critical challenge is that they may not proactively determine when or how to apply different thinking processes.
Instead, their thinking process is often shaped by pattern matching from training data and often diverges from human reasoning patterns~\cite{dziri2024faith,bao-etal-2025-likely}, rather than selecting the most appropriate way to think or respond.

To address these challenges, researchers have sought to explicitly define the thinking processes of LLMs by mimicking aspects of human cognition and introducing reasoning strategies such as self-verification~\cite{weng2022large,madaan2024self}, chain-of-thought prompting~\cite{wei2022chain}, and reverse thinking~\cite{chen2024reverse}, etc.
By constraining LLMs to follow these structured thinking processes, models have shown improved performance on specific tasks.
However, these methods are often designed for particular types of tasks, limiting their adaptability and effectiveness across diverse tasks~\cite{sprague2024cot,xu-etal-2024-pride}.
More critically, these approaches impose fixed thinking structures rather than enabling LLMs to adaptively determine the most effective task-specific strategy, potentially limiting their performance.

To address these limitations, we propose the idea of \emph{meta-thinking}, a process where LLMs first reflect on their approach before generating a response.
Rather than immediately solving a problem, the model initiates the process by determining how to think and select the most suitable cognitive strategy from a range of available options.
For example, when tackling a complex logical puzzle, an LLM might choose to work forward from given conditions, reason backward from the goal, or verify its answer through self-reflection.
By incorporating this meta-thinking step, LLMs can dynamically adapt their reasoning process to different tasks, rather than relying on rigid, predefined heuristics.

\begin{figure*}[t]
    \centering
    \includegraphics[width=\linewidth]{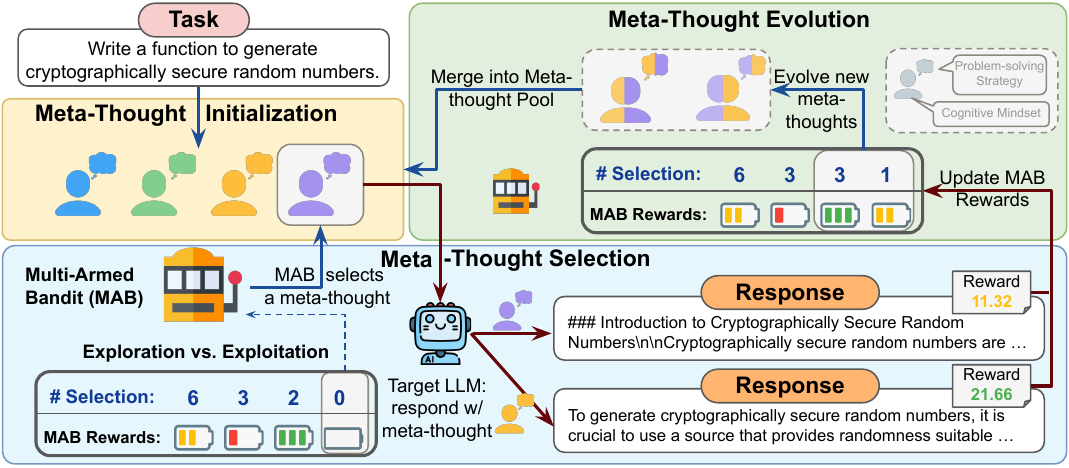}
    \caption{Overview of \MODEL. \MODEL initializes a pool of meta-thoughts, selects and evaluates them using the Upper Confidence Bound (UCB) algorithm, and periodically evolves high-performing meta-thoughts. The best response is returned after exhausting the sampling budgets.}
    \label{fig:main}
\end{figure*}

Building on this idea, we introduce \MODEL, a test-time scaling framework that incorporates meta-thinking. The goal is to enable LLMs to explore different thinking strategies, and generate the most effective response for a given input.
The process begins with the generation of an initial pool of meta-thoughts, where the LLM autonomously formulates reasoning strategies tailored to the given problem. To enhance diversity, we retrieve relevant input-output pairs from instruction-tuning datasets~\cite{zhao2024wildchat} and prompt the LLM to abstract thinking patterns specific to the current task.
Next, we iteratively prompt the LLM based on the selected meta-thoughts.
The selection process is guided by an Upper Confidence Bound (UCB; \citealt{auer2002using}) algorithm, which balances exploration and exploitation. A reward model~\cite{liu2024skywork} then evaluates response quality. To further promote diversity, we periodically apply a \emph{genetic algorithm}, generating new meta-thoughts by evolving high-reward ones.
Finally, the best response found through this process is returned after reaching the sampling budget.

Experiments show that \MODEL significantly enhances the problem-solving capabilities of LLMs across diverse tasks, consistently outperforming baseline methods on both various real-world reasoning challenges. More importantly, \MODEL scales effectively with increased sampling budgets, further improving performance by evolving meta-thoughts that leads to task solutions of higher quality.

Our contributions are three-fold. First, we introduce \MODEL, a novel test-time scaling strategy that enables LLMs to proactively select and refine cognitive strategies at test time, moving beyond static, predefined heuristics. Second, we adapt the genetic algorithm to evolve high-reward meta-thoughts over multiple iterations, further enhancing adaptability and reasoning efficiency. Third, experiments demonstrate that \MODEL consistently scales with increased sampling budgets, effectively leveraging additional computational resources to refine responses.







\section{Method}
\label{method}
In this section, we introduce \MODEL, a test-time scaling framework that explores diverse cognitive strategies to optimize the reasoning process during inference. 
We first provide a brief overview and relevant preliminaries in \Cref{method:overview}, followed by a detailed description of meta-thought initialization (\Cref{method:init}), selection (\Cref{method:selection}), and evolution (\Cref{method:evolution}).

\subsection{\MODEL Overview}
\label{method:overview}

\MODEL is a test-time scaling strategy designed to enhance the problem-solving capabilities of LLMs by incorporating \emph{meta-thoughts} that guide the reasoning process before generating a response. Inspired by human cognitive processes, a \emph{meta-thought} consists of two key components: 
\begin{enumerate}
\setlength\itemsep{0em}
    \item \textbf{Cognitive Mindset}, which is an appropriate perspective, expertise, or role that the model adopts to approach the task.
    \item \textbf{Problem-Solving Strategy}, which is a structured pattern used to formulate a sound solution based on the given mindset.
\end{enumerate}
For instance, when asked to ``\emph{Use ABC notation to write a melody in the style of a folk tune}'', a well-formed meta-thought could include a cognitive mindset of ``\emph{a musician or music educator with expertise in music theory and composition}'', and a problem-solving strategy indicating ``\emph{First present a melody that captures the characteristics of a traditional folk tune, then \dots}''
The model then utilizes these strategies to guide its reasoning process and generate a response.

\MODEL develops effective meta-thoughts for an input task in three phases: initialization, selection, and evolution. 
During initialization, \MODEL generates a diverse pool of reasoning strategies tailored to the input task by leveraging its own prior knowledge and instruction-tuning datasets. Next, a Multi-Armed Bandit (MAB) algorithm is used to explore and select promising meta-thoughts. 
Finally, \MODEL employs a genetic algorithm to iteratively refine and expand the meta-thought pool. To maintain computational feasibility, \MODEL operates within a fixed sampling budget, and returns the response with the highest score given by a reward model once the budget is exhausted. We illustrate the overall design of \MODEL in \Cref{fig:main}.





Before delving into the details of \MODEL, we provide the background knowledge on genetic algorithms and multi-armed bandit (MAB) algorithms, two key components of our framework. Genetic algorithms enable iterative refinement of meta-thoughts, while MAB facilitates selection of the most effective meta-thought during inference.

\stitle{Genetic Algorithms}
Genetic algorithms~\cite{golberg1989genetic,mitchell1998introduction} are optimization techniques inspired by natural selection. They consist of three key components: \emph{initialization}, which generates an initial population of candidate solutions; \emph{fitness evaluation}, which assesses solution quality based on a predefined objective; and \emph{evolution}, where high-fitness candidates are selected (\emph{selection}), combined (\emph{crossover}), or modified (\emph{mutation}) to create new solutions.
Through iterative refinement, the population converges toward solutions with better fitness. The process terminates once a predefined fitness threshold is reached or a set number of generations is completed.

\stitle{Multi-Armed Bandit}
The MAB problem addresses the trade-off between exploration (testing new options) and exploitation (choosing known high-reward options) in sequential decision-making~\cite{vermorel2005multi,audibert2009exploration}. The goal is to maximize cumulative rewards by selecting from multiple options (arms). A key approach, the Upper Confidence Bound (UCB) algorithm~\cite{auer2002finite,auer2002using}, balances this trade-off by combining an option’s empirical mean reward with an uncertainty bonus that decreases with more sampling. This ensures efficient exploration while favoring consistently high-performing choices.

\subsection{Meta-Thought Initialization}
\label{method:init}

The initialization of meta-thoughts in \MODEL aims to generate a diverse set of reasoning strategies tailored to the given problem. This process follows two complementary strategies.
First, the target LLM is prompted to self-compose a set of reasoning strategies for the given problem, reflecting different problem-solving heuristics and cognitive patterns that can enhance inference performance.
Second, to further enrich diversity, \MODEL leverages instruction-tuning datasets such as WildChat~\cite{zhao2024wildchat}, which contain extensive task-solution pairs generated by advanced models. We retrieve top $8$ similar tasks (measured by the similarity between the embeddings of task descriptions) from these datasets and prompt the LLM to extract and adapt the underlying thinking patterns to the current problem.\footnote{The prompts used for this process are illustrated in \Cref{append:prompt}.} The assumption is that reasoning strategies used for similar questions can offer valuable perspectives for solving new ones.
Together, these self-composed and dataset-derived strategies form the initial pool of meta-thoughts used to guide problem-solving.

\subsection{Meta-Thought Selection as MAB}
\label{method:selection}
Once the meta-thought pool is initialized, \MODEL selects the most promising meta-thought for generating a response at each iteration step. We here formalize the selection of meta-thoughts at inference time as a multi-armed bandit (MAB) problem, where each meta-thought option serves as an ``arm'' and the goal is to maximize the reward of the corresponding model response based on the selected meta-thought.

Specifically, we apply the Upper Confidence Bound (UCB) algorithm to guide the selection process by balancing exploration and exploitation of the available meta-thoughts, which dynamically selects a candidate based on both its past performance and the uncertainty about its future performance. Specifically, given a candidate set of meta-thoughts $\mathcal{M}_q$ initialized for an input query $q$, the selected meta-thought $M^*$ is determined by 
\begin{equation*}
M^* = \arg\max_{M \in \mathcal{M}_q} \left( \mu_M + \beta \sqrt{\frac{\log t}{N_M}} \right),
\end{equation*}
where $\mu_M$ represents the empirical mean reward of meta-thought $M$, computed based on past attempts, while $N_M$ denotes the number of times $M$ has been selected. $\beta \sqrt{\frac{\log t}{N_M}}$ introduces an exploration term that prioritizes meta-thoughts that have been less frequently tested, where $t$ is the total number of attempts processed. The parameter $\beta$ controls the trade-off between exploiting well-performing options and exploring new ones.

To assess response quality during fitness evaluation, we use an outcome-based general-purpose reward model~\cite{liu2024skywork}. Given the model input and response, the reward model evaluates response quality and assigns a reward score.
This reward score, in turn, updates the UCB of the meta-thought, which influences subsequent selection decisions by refining the exploration-exploitation balance according to the accumulated performance history.

\subsection{Genetic Meta-Thought Evolution}
\label{method:evolution}
To further enhance diversity and continuously expand the pool of cognitive strategies, \MODEL periodically applies a genetic algorithm to evolve the meta-thoughts. High-performing meta-thoughts, identified by their accumulated UCB as described in \Cref{method:selection}, are selected as parents for producing the next generation of new child meta-thoughts. Specifically, given the high-UCB meta-thoughts, the target LLM is prompted to develop a set of refined meta-thoughts that integrates and improves upon the selected parents.

Rather than directly applying explicit crossover or mutation policies as usually applied in a typical genetic algorithm, we prompt the target LLM itself to determine how to combine the reasoning heuristics of the parent meta-thoughts and evolve them into improved counterparts.
We find that directly performing crossover or mutation at the text level often produces strategies that make no sense.
Instead, prompting LLMs allows for a more flexible and context-sensitive evolution of meta-thoughts, ensuring that the evolved strategies continue to reflect the problem-solving patterns most effective for the current task over successive iterations.

Based on the test-time scaling nature of \MODEL, we set the termination criterion according to a predetermined budget for the number of attempts. Once the budget is exhausted, \MODEL terminates and returns the model response (which is prompted with meta-thought) with the highest reward score among all the attempts as evaluated by the reward model. The overall algorithm of \MODEL is formally outlined in \Cref{alg:meta_scale}.

\begin{algorithm}[t]
\caption{\MODEL}
\label{alg:meta_scale}
\begin{algorithmic}[1]
\Require Query $q$, target LLM, reward model \textbf{RM}, instruction-tuning dataset $\mathcal{D}$, budget $T$, evolution interval $k$
\Ensure Optimized response $\hat{y}$
\State \textbf{Initialize} meta-thought pool $\mathcal{M}$ (\Cref{method:init})

\For{$t = 1$ to $T$}
    \State Select meta-thought $M^*$ using UCB (\Cref{alg:meta_scale}, \Cref{method:selection})
    \State Generate response $y_t$ with $M^*$
    \State Compute reward score $\textbf{RM}(y_t)$
    \State Update UBC for $M^*$ and record reward score for $y_t$
    
    \If{$t \, \% \, k == 0$}
        \State Select top-performing meta-thoughts $\mathcal{M}_{\text{top}} \subset \mathcal{M}$
        \State Generate new meta-thoughts $\mathcal{M}_{\text{new}}$ using LLM (\Cref{method:evolution})
        \State Merge $\mathcal{M}_{\text{new}}$ and $\mathcal{M}$
    \EndIf
\EndFor
\State Return best response $\hat{y} = \arg\max_{y_t} \textbf{RM}(y_t)$
\end{algorithmic}
\end{algorithm}

\begin{table*}[t]
\centering
\begin{tabular}{lcccccc}
\toprule
\multirow{3}{*}{\textbf{Method}} & \multicolumn{4}{c}{\textbf{Arena-hard}} & \textbf{MMLU-Pro} & \textbf{GSM8K} \\ \cmidrule(lr){2-5}\cmidrule(lr){6-6}\cmidrule{7-7}
 & Avg. & 95\% CI & 95\% CI & Number of & Acc  & Acc \\
& Win Rate & Lower& Upper& Tokens&& \\
 \midrule
\multicolumn{7}{c}{GPT-4o-0806} \\
\midrule
Vanilla Model 1-Pass & 82.14 & 77.68 & 88.10 & 573 & 68.75 & 92.19 \\
Chain-of-Thought 1-Pass & 84.72 & 79.30 & 88.91 & 663 & 71.88 & 93.75 \\ \cdashline{1-7}
Best-of-N & 84.17 & 78.73 & 89.31 & 594 & 71.88 & 92.19 \\
Best-of-N w/ CoT & 87.42 & 81.66 & 91.37 & 681 & 73.44 & \textbf{95.31} \\ \cdashline{1-7}
\MODEL w/o Evolution & 89.63 & 84.57 & 93.62 & 689 & 76.56 & 93.75 \\
\MODEL & \textbf{93.14} & \textbf{90.44} & \textbf{96.34} & 699 & \textbf{78.12} & \textbf{95.31} \\ \midrule
\multicolumn{7}{c}{Llama-3.1-8B-Instruct} \\ \midrule
Vanilla Model 1-Pass & 18.94 & 13.50 & 23.50 & 481 & 31.25 & 73.44 \\
Chain-of-Thought 1-Pass & 12.50 & 8.74 & 17.50 & 625 & 35.94 & 78.13 \\ \cdashline{1-7}
Best-of-N & 25.88 & 16.76 & 32.37 & 493 & 39.06 & 79.69 \\
Best-of-N w/ CoT & 28.33 & 21.63 & 34.74 & 571 & 45.31 & 82.81 \\ \cdashline{1-7}
\MODEL w/o Evolution & 28.93 & 23.02 & 35.26 & 575 & 46.88 & \textbf{84.38} \\
\MODEL & \textbf{30.86} & \textbf{23.97} & \textbf{35.29} & 630 & \textbf{50.00} & 82.81 \\ \bottomrule
\end{tabular}
\caption{Main results (\%) on Arena-Hard, MMLU-Pro, and GSM8K, evaluated using GPT-4o and Llama-3.1-8B-Instruct as base models. The best scores for each base model are highlighted in \textbf{bold}.}
\label{tab:main}
\end{table*}

\section{Experiments}
\label{sec:exp}

In this section, we evaluate \MODEL on three different problem-solving tasks with two models of varying scales. We provide an overview of our experimental settings (\Cref{exp:setup}) and present a comparison of empirical results (\Cref{exp:main_results}) followed by further analysis (\Cref{exp:analysis}).

\subsection{Experimental Setup}
\label{exp:setup}

\paragraph{Evaluation Dataset}
To evaluate the effectiveness of our \MODEL for problem-solving, we consider the following mathematical reasoning benchmark and general-purpose tasks.
~\\\textbf{GSM8K}~\citep{cobbe2021training} consists of high-quality, linguistically diverse grade school math word problems.
We use the accuracy to measure the models' performance.
To compute accuracy, we parse the completion to extract the numeric value and conduct an exact match with the reference answer.
~\\\textbf{MMLU-Pro}~\citep{wang2024mmlu} is a 10-way multiple-choice extended version of the MMLU dataset~\citep{hendrycks2020measuring}. Spanning 14 diverse domains, we use it to evaluate the proficient-level multi-discipline language understanding and reasoning capabilities of LLMs. To extract option letter from responses, we employ the suggested regular expression matching (i.e., \texttt{`answer is \textbackslash(?\textbackslash([A-J]\textbackslash)?\textbackslash)'}) first, followed by the secondary regex matching (i.e., \texttt{`\textbackslash.*\textbackslash[aA\textbackslash]nswer:\textbackslash s*\textbackslash([A-J]\textbackslash)'}). If both matching methods fail to retrieve a valid response, a random option from the answer choices is selected. The correctness averaged over all testing cases is used as the final accuracy.
~\\\textbf{Arena-Hard}~\citep{li2024crowdsourced} is composed of challenging benchmark prompts sourced from Chatbot Arena~\citep{chiang2024chatbot} and adopts the LLM-as-a-Judge for automatic model evaluation. We follow their default evaluation setting by using GPT-4-Turbo as the judge to estimate human preferences of evaluated models against the baseline model  GPT-4-0314, and report the average win rate.
Additionally, following the official evaluation setting, we report the 95\% confidence interval of the win rate, providing both the lower and upper bounds. We also report the number of tokens in the generated responses.

\paragraph{Baseline Methods}
\label{exp:baseline}

For each task, we compare \MODEL to four baseline inference methods with the same reward model and scaling budget if applicable.
~\\\textbf{1-Pass}, where the target LLM performs a single forward pass with greedy decoding where the temperature is fixed to $0$.
~\\\textbf{Chain-of-Thought} (CoT, \citealt{wei2022chain}), where the target LLM is prompted to generate a reasoning process towards the final solution for the given task. This is also performed in a greedy decoding manner with the temperature set as $0$.
~\\\textbf{Best-of-N}~\cite{brown2024large}, where up to $128$ candidate solutions are sampled with temperature of $0.6$ that allows diversity. The solution with the highest score given by the reward model is returned as the final response.
~\\\textbf{Best-of-N with CoT}, where the sampling budget is also $128$ and the LLM is prompted to perform CoT for each sampling. Temperature of $0.6$ is applied.

\paragraph{Models}
We use one state-of-the-art LLM, GPT-4o (gpt-4o-0806)~\cite{achiam2023gpt}, and one open-source LLM Llama-3.1-8B-Instruct~\cite{dubey2024llama} for evaluation. Skywork-Reward-Llama-3.1-8B-v0.2~\cite{liu2024skywork} is directly adopted as the reward model for assigning reward scores to candidate responses, based on which the optimal response is chosen as the final answer. To ensure fair comparisons, we use consistent inference settings across models and maintain the same sampling budgets.

\paragraph{Implementation Details}
\label{exp:imp}
Due to limited computational resources and high API costs, we randomly sample $64$ testing instances from each benchmark to create new subsampled evaluation sets, which are used consistently across all methods. To precisely evaluate the effectiveness of different test-time compute scaling strategies without interference from demonstrations, we prompt LLMs in the zero-shot setting on all studied tasks.

\begin{table}[t]
\resizebox{\columnwidth}{!}{
\begin{tabular}{lcccc}
\toprule
\textbf{Model} & \textbf{Avg.} & \textbf{95\% CI} & \textbf{95\% CI} & \textbf{Number of} \\ 
& \textbf{Win Rate} & \textbf{Lower} & \textbf{Upper} & \textbf{Tokens} \\ \midrule
claude-3-5-sonnet & 82.0 & 80.4 & 84.2 & 567 \\
o1-preview-2024-09-12 & 81.6 & 79.2 & 83.8 & 1193 \\
\rowcolor[HTML]{ECF4FF} 
$\MODEL_{128}$ & 80.1 & 78.2 & 82.2 & 744 \\
o1-mini-2024-09-12 & 79.2 & 76.6 & 81.6 & 1399 \\
\rowcolor[HTML]{ECF4FF} 
$\MODEL_{64}$ & 78.8 & 76.1 & 81.5 & 737 \\
\rowcolor[HTML]{ECF4FF} 
$\MODEL_{32}$ & 76.7 & 74.4 & 79.9 & 739 \\
\rowcolor[HTML]{ECF4FF} 
$\MODEL_{16}$ & 74.7 & 71.8 & 77.4 & 735 \\
gpt-4-turbo-2024-04-09 & 74.4 & 71.9 & 76.5 & 662 \\
gpt-4-0125-preview & 73.5 & 71.1 & 75.3 & 619 \\
\rowcolor[HTML]{ECF4FF} 
$\MODEL_{8}$ & 71.4 & 68.7 & 74.2 & 734 \\
gpt-4o-2024-08-06 & 71.0 & 68.5 & 73.8 & 594 \\ \bottomrule
\end{tabular}
}
\caption{Results on full Arena-Hard dataset under style control \cite{chiang2024chatbot} for a fair comparison that eliminates the effects of style (e.g. answer token length and number of markdown headers) on response judgment. $\MODEL_{k}$ represents \MODEL based on gpt-4o-2024-08-06 with sampling budget of $k$. The rest of the models perform a single forward pass. Evaluation results without style control are shown in \Cref{exp:arena_wo}.}
\label{exp:arena_style}
\end{table}

\subsection{Main Results}
\label{exp:main_results}
\stitle{\MODEL Consistently Outperforms Baselines}  
We present the main results in~\Cref{tab:main} and find that \MODEL consistently achieves equal or superior performance compared to both one-pass methods and Best-of-N methods, regardless of whether they are prompted with or without CoT.  
Notably, \MODEL based on GPT-4o outperforms its one-pass counterpart by 11.00\%, 9.37\%, and 3.12\% on Arena-Hard, MMLU-Pro, and GSM8K, respectively. It also surpasses the standard Best-of-N method by 8.97\%, 6.24\%, and 3.12\%, respectively. Further, GPT-4o outperforms o1-mini \cite{jaech2024openai} with \MODEL under style control (\Cref{exp:arena_style}).
Similar improvements are observed when using Llama-3.1-8B-Instrct as the base model.  
These results demonstrate that integrating meta-thoughts enables LLMs to scale more effectively during test time as the number of samples increases.

\stitle{Evolving Meta-Thoughts Improves Performance}
We compare the performance of \MODEL with and without the evolution of meta-thoughts during testing. Our results show that \MODEL achieves better performance across most evaluation datasets and models, with only a slight underperformance on GSM8K when using Llama-3.1-8B-Instruct. This demonstrates that evolving meta-thoughts during test time leads to more effective reasoning and decision-making.

\begin{figure}[t]
    \centering
    \includegraphics[width=0.48\textwidth]{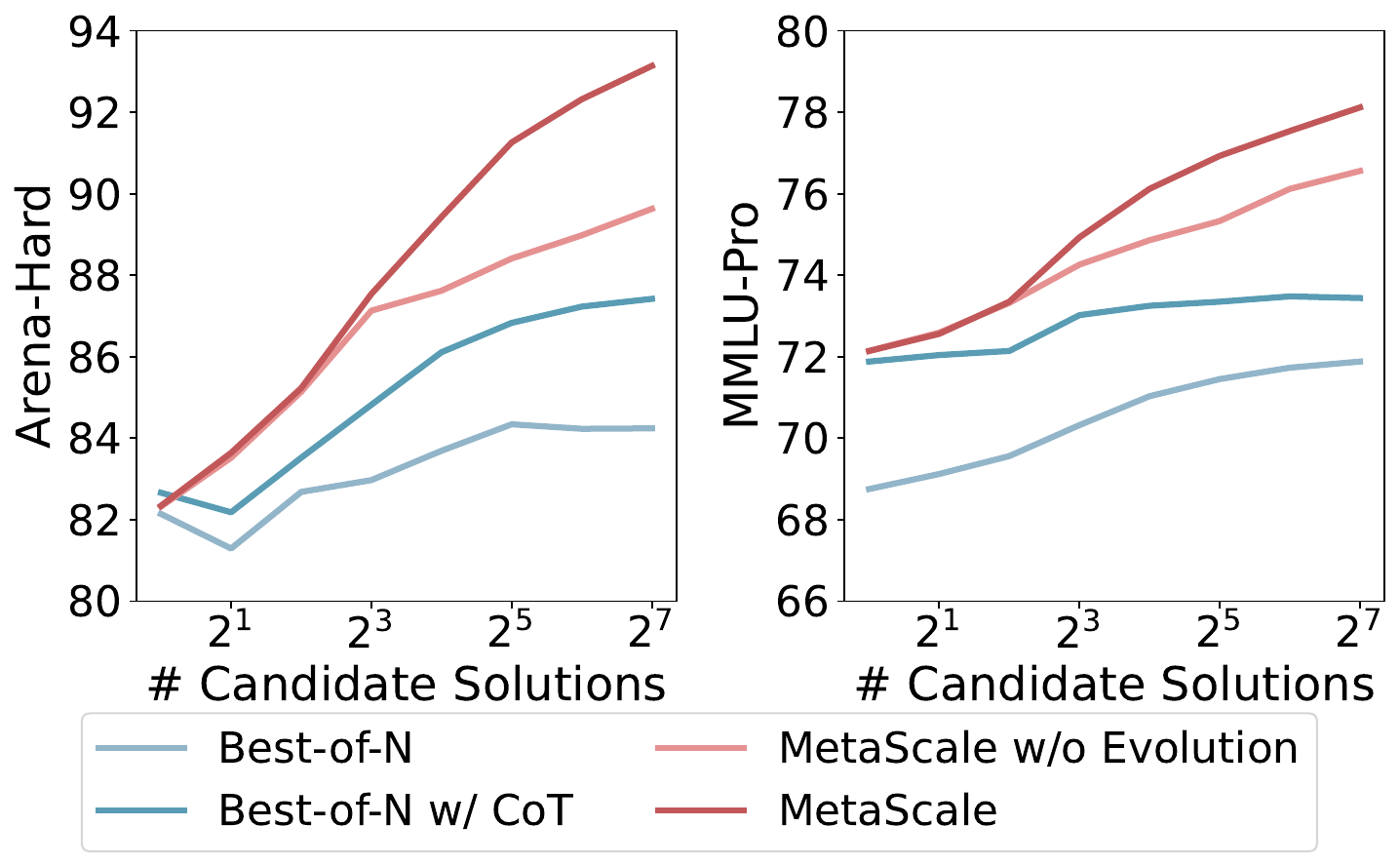}
    \caption{Results (\%) on Arena-Hard and MMLU-Pro with varying sampling budgets (1 to 128) using GPT-4o as the base model. \MODEL improves more rapidly as the sampling budget increases, whereas best-of-N shows slower gains or even stops to increase at the maximum budget.}
    \label{fig:scale}
\end{figure}

\subsection{Analysis}
\label{exp:analysis}

\stitle{\MODEL Benefits from Increased Sampling Budgets}
As illustrated in \Cref{fig:scale}, \MODEL demonstrates significantly higher gains as the number of candidate solutions increases compared to baseline methods. On both Arena-Hard and MMLU-Pro benchmarks, \MODEL consistently outperforms other test-time scaling methods, i.e. Best-of-N and Best-of-N w/ CoT, with its performance improving more steeply as the sampling budget expands. This suggests that \MODEL is a more effective scaling strategy that efficiently explores better solutions, whereas Best-of-N slows down at higher budgets. The evolutionary nature of \MODEL further amplifies this advantage, enabling the target LLM to apply refined heuristics through iterative meta-thought selection and evolution.

\begin{figure}[t]
    \centering
    \includegraphics[width=0.48\textwidth]{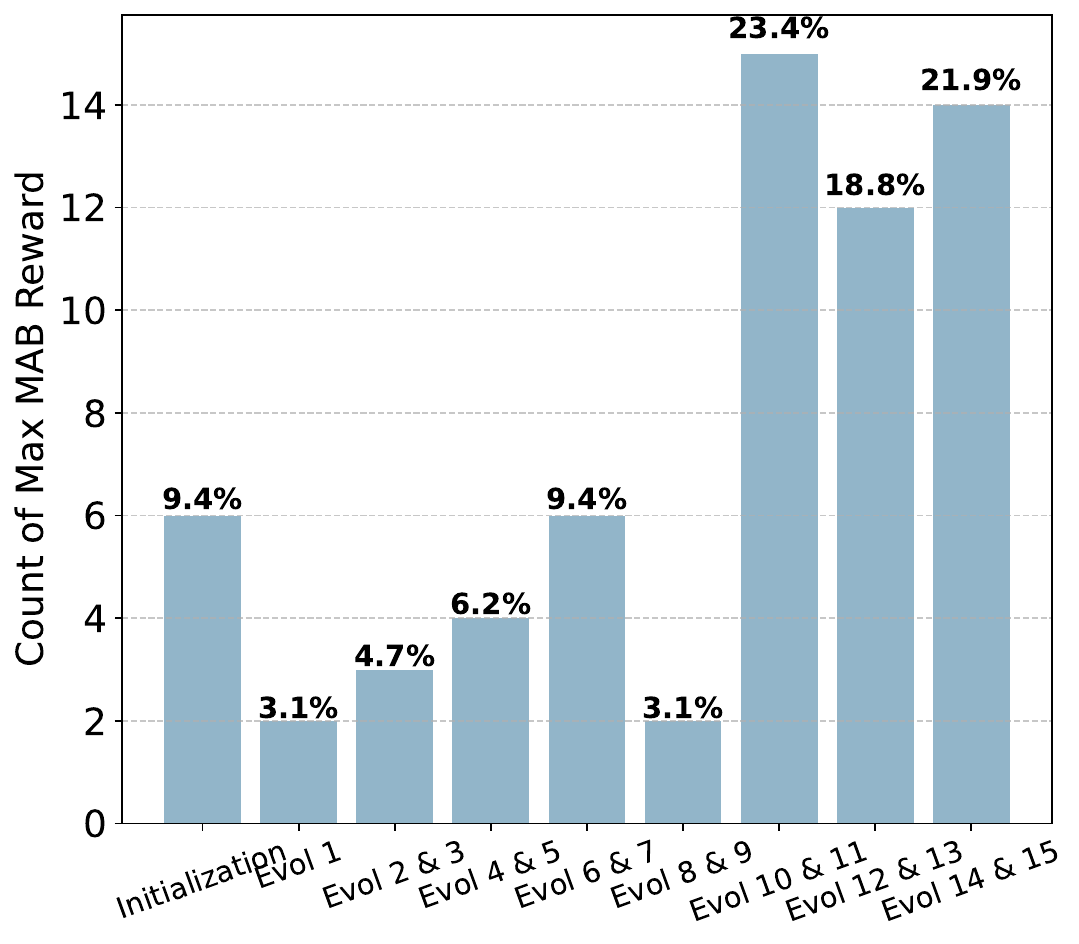}
    \caption{Distribution of selected meta-thoughts after the full \MODEL process on Arena-Hard using GPT-4o.}
    \label{fig:MAB}
\end{figure}

\stitle{Evolutionary Refinement Enhances Selection of Optimal Meta-thoughts}
\Cref{fig:MAB} demonstrates how the distribution of selected meta-thoughts shifts over multiple evolutionary iterations throughout the \MODEL process. Notably, the frequency of highly rewarded (named MAB reward) meta-thoughts increases significantly as the iteration of meta-thought evolution proceeds, particularly from the tenth iteration onward, where a substantial portion of selections of over $20\%$ occur. This suggests that the iterative meta-thought refinement process within \MODEL enables the model to progressively converge on more optimal solutions. The lower selection counts in earlier stages indicate a wider exploration phase before \MODEL effectively approaches the best-performing meta-thoughts. Overall, the evolutionary refinement of meta-thoughts allows for better performance indicated by higher chances of hitting the maximum MAB reward.

\begin{figure*}[t]
    \centering
    \includegraphics[width=\linewidth]{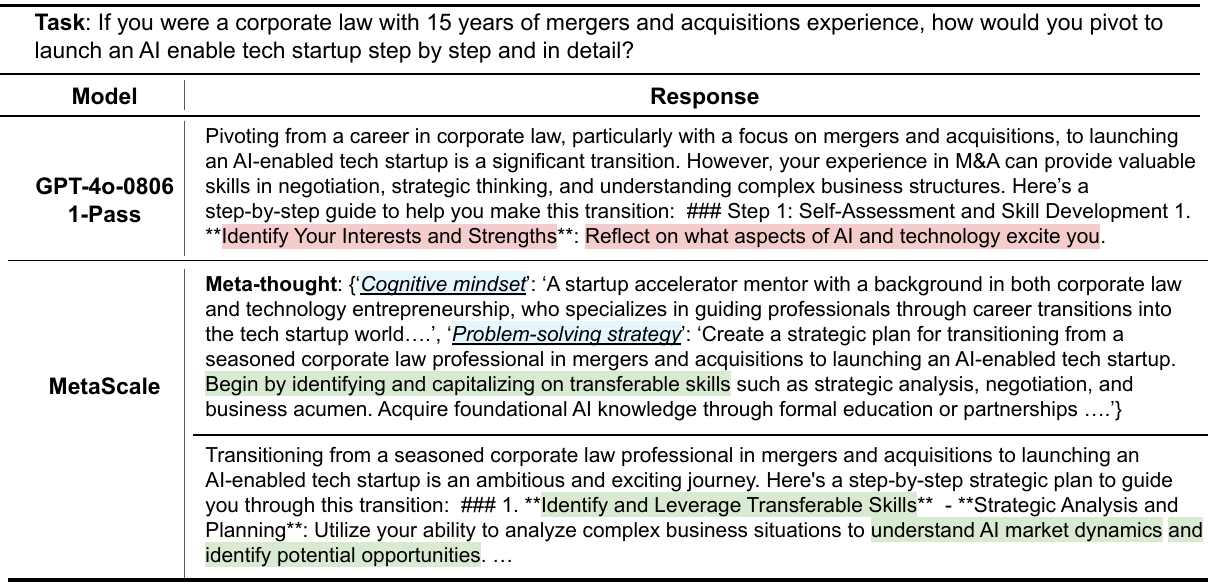}
    \caption{Model responses for a task from Arena-hard benchmark. The meta-thought used for the optimal response which obtains the highest reward among all candidate solutions sampled by \MODEL is marked \blue{in blue}. Based on the optimized heuristic provided by the meta-thought, \MODEL tackles the task in a more sophisticated way that \green{precisely targets the core challenges} of the problem, hitting a solution with maximal efficiency and quality. In contrast, the vanilla GPT-4o-0806 could only provide \pink{general ideas} which may lack professional insights.}
    \label{fig:case_study}
\end{figure*}

\stitle{\MODEL Generates Targeted and Expert-Level Solutions}
We provide a case study of model responses in \Cref{fig:case_study}, comparing a standard GPT-4o single-pass response with an optimized response generated by \MODEL. Given the task of launching a startup based on past experience, the single-pass response offers broad and generic guidance, failing to identify the core challenge of transitioning from corporate law to an AI startup. With a lack of specificity, the LLM provides surface-level advice without deep insights into the strategic and operational challenges involved.

In contrast, \MODEL refines its response by incorporating iteratively refined meta-thoughts, which allow for a more structured, expert-driven approach. The optimized response precisely identifies key transferable skills, highlights AI market dynamics, and presents a clear, step-by-step strategy tailored to the user's professional background. By leveraging meta-thought, the response is not only contextually relevant but also highly aligned with domain-specific expertise.

\section{Related Work}
\stitle{Test-time Compute Scaling}
As a complementary approach to prohibitively expensive train-time compute scaling (i.e., larger models pre-trained with growing amounts of tokens), test-time compute scaling has exhibited preliminary success on both proprietary~\citep{snell2024scaling} and open-source models~\citep{HuggingFaceH4ScalingTestTimeCompute}. There are two major ways to scale test-time compute: \emph{1) self-refinement} where models iteratively generate and revise responses based on internal or external feedback~\citep{xi2024enhancing,song2024mind,cheng2024spar}, and \emph{2) searching against a verifier} where models sample multiple candidate answers and a verifier is used to select the best one~\citep{huang2024deal,wang2024openr,guan2025rstar}. However, self-refinement relies on built-in correction capabilities from the generative model, which is not sufficiently emphasized during model training~\citep{ji2025test}. Empirical studies on tasks such as code generation~\citep{olausson2023self} and commonsense QA~\citep{huang2023large} show that self-refinement is not a guaranteed solution for performance improvement. On the other hand, searching-based strategies lead to performance gain with a flexible verifier design, ranging from hard-coded heuristics~\citep{snell2024scaling} to learned process- or outcome-based reward models~\cite{HuggingFaceH4ScalingTestTimeCompute}. Searching approaches such as Best-of-N sampling~\citep{cobbe2021training,lightman2023let}, beam search~\citep{feng2023alphazero,yao2023tree} and Monte Carlo tree search (MCTS;~\citealt{tang2024dawn,zhang2024accessing}) mainly demonstrate benefits for enhancing performance on tasks requiring extensive reasoning such as mathematical (e.g., GSM8k;~\citealt{cobbe2021training}) and logical reasoning (e.g., PrOntoQA;~\citealt{saparov2022language}). Orthogonal to prior test-time compute scaling methods, our proposed \MODEL searches for the best thinking process, which more effectively elicits capabilities from LLMs to handle both reasoning and general-purpose tasks.

\stitle{Persona-assigned LLMs}
By instructing LLMs to mimic a wide range of personas (e.g., demographic, character and individualized personas), we have observed their improved social intelligence and theory of mind~\citep{sap2022neural,kosinski2024evaluating}. Besides, recent work~\citep{xu2023expertprompting,wang2023unleashing} has shown that LLMs or agents assigned with specific personas are able to unleash the potential of cognitive synergy in LLMs, which improves their problem-solving abilities compared to using the default ``helpful assistant'' persona. Moreover, the persona-driven methodology~\citep{ge2024scaling} has become an efficient way to construct synthetic pre-training (e.g., TuluMath for OLMo 2 training;~\citealt{olmo20242}) and post-training (e.g., Tülu-3-Persona-IF for Tülu 3 training;~\citealt{lambert2024t}) data. Going beyond persona-assigned LLMs, we further unleash their potential for solving challenging problems by automatically retrieving relevant meta-thoughts and prompting LLMs with evolved theory of mind.

\stitle{Cognitive Strategies for Problem Solving}
To improve performance on reasoning and planning benchmarks (e.g., GSM8K;~\citealt{cobbe2021training} and BigBench;~\citealt{srivastava2022beyond}),  various methods have been proposed to induce specific reasoning structures mimicking the reasoning structure of the underlying task associated with the dataset, e.g., chain-of-thought~\citep{wei2022chain}, question summarization~\citep{kuznia-etal-2022-less} and question decomposition~\citep{patel-etal-2022-question}, etc. To address real-world queries covering diverse reasoning structures, recent work proposes to uncover the intrinsic reasoning structure for solving studied task via meta reasoning, where LLMs are instructed to select, adapt and implement actionable reasoning~\citep{zhou2024self,gao2024meta}. However, existing meta-reasoning methods rely heavily on pre-defined cognitive heuristics of problem-solving (e.g., ``Use critical thinking'' and ``Let’s think step by step'';~\citealt{fernando2023promptbreeder}), which is of limited coverage hence provides suboptimal solutions for tasks requiring rarely-seen or complex skills. To broaden the coverage of useful cognitive structures for problem solving, \MODEL retrieves from massive high-quality real-world conversations between users and assistants and extracts relevant meta thoughts automatically. Moreover, \MODEL incorporates an effective genetic algorithm to further promote diversity.



\section{Conclusion}
We introduced \MODEL, a test-time scaling framework that enables LLMs to adaptively select and refine cognitive strategies, namely meta-thoughts, through iterations. By leveraging a multi-armed bandit algorithm for strategy selection and a genetic algorithm for iterative refinement, \MODEL dynamically optimizes reasoning processes, scaling effectively with increased sampling budgets. Experiments on GSM8K, MMLU-Pro, and Arena-Hard show that \MODEL consistently outperforms baselines, achieving an absolute 11\% win-rate improvement on Arena-Hard with GPT-4o. These results highlight meta-thinking as a promising approach for enhancing LLM reasoning.

\section*{Limitations}
In this work, we propose the cognitive strategy, \MODEL, to more effectively solve challenging problems requiring diverse capabilities. The limitations of this work are as follows: \emph{1) Language coverage}: we mainly focus on tasks represented in English, hence meta thoughts are extracted from English-based conversations so that final responses are expressed in English. Motivated by the observed performance improvement on English benchmarks, we expect similar performance gain from \MODEL on other language-based tasks. \emph{2) Collaborative LLMs}: we show effectiveness of \MODEL on single LLMs. Considering the diversity of inherent cognition and reaction towards meta-thought prompting among LLMs, we believe further performance improvement can be achieved by applying \MODEL to LLMs from different families and of distinct sizes and incorporating their responses.




\section*{Ethics Statement}
This paper presents comprehensive study of meta-thoughts in challenging problem-solving scenarios.  Both the utilized proprietary and open-source LLMs have gone through thorough safety alignment and evaluation by model developers. Meanwhile, data sources for meta-thought extraction and evaluation benchmarks are publicly available with required ethical reviews conducted in prior work. Therefore, we believe our work does not pose additional ethical issues.



\bibliography{anthology,custom}
\bibliographystyle{acl_natbib}

\appendix


\newpage
\begin{center}
    {\Large\textbf{Appendices}}
\end{center}

\section{Prompts for Meta-Thought Initialization and Evolution}
\label{append:prompt}

\begin{tcolorbox}[colback=white, colframe=gray!80!gray, boxrule=0.5pt, sharp corners, title=Prompt for Self-Compose]
\small
Given the following question, 1) who is likely to give appropriate answer? please provide a concise description of the persona, 2) provide a high-level abstract of how the person will answer the question reasonably, do not discuss any details in the question. 
\newline

Question: Old age PT hx of DM, HTN, dyslipidemia His ECG I.II, aVF (MI) what is the highest risk factor for this condition?\\
Persona: a cardiologist—a medical doctor specializing in diagnosing and treating diseases of the cardiovascular system\\
High-level abstract: first assess the patient's various risk factors, considering their relative contributions to the development of the condition; then identify which risk factor poses the highest risk for the patient's condition by evaluating the impact of each factor based on medical knowledge and epidemiological data, the cardiologist
\newline

Question: Which singer is better technically: Floor Jansen or Taylor Swift? Rate from 1 to 10 your confidence that your answer is correct.\\
Persona: a professional vocal coach with extensive experience in assessing singers' technical abilities across various music genres\\
High-level abstract: assess and compare the technical abilities of both singers and then determine who is technically better, ending with a confidence rating

\end{tcolorbox}

\section{Results on Arena-Hard}

\begin{table}[h]
\resizebox{\columnwidth}{!}{
\begin{tabular}{lcccc}
\toprule
\textbf{Model} & \textbf{Avg.} & \textbf{95\% CI} & \textbf{95\% CI} & \textbf{Number of} \\
& \textbf{Win Rate} & \textbf{Lower} & \textbf{Upper} & \textbf{Tokens} \\ \midrule
o1-mini-2024-09-12 & 92.0 & 90.8 & 93.0 & 1399 \\
o1-preview-2024-09-12 & 90.4 & 89.3 & 91.7 & 1193 \\
\rowcolor[HTML]{ECF4FF} 
$\MODEL_{128}$ & 89.0 & 87.5 & 90.4 & 744 \\
\rowcolor[HTML]{ECF4FF} 
$\MODEL_{64}$ & 88.2 & 86.7 & 89.5 & 737 \\
\rowcolor[HTML]{ECF4FF} 
$\MODEL_{32}$ & 86.9 & 85.5 & 88.5 & 739 \\
\rowcolor[HTML]{ECF4FF} 
$\MODEL_{16}$ & 85.3 & 83.3 & 87.0 & 735 \\
llama-3.1-70b & 84.9 & 83.2 & 86.7 & 869 \\
\rowcolor[HTML]{ECF4FF} 
$\MODEL_{8}$ & 82.8 & 80.7 & 84.3 & 734 \\
\rowcolor[HTML]{ECF4FF} 
$\MODEL_{4}$ & 82.8 & 80.9 & 85.0 & 728 \\
gpt-4-turbo-2024-04-09 & 82.6 & 80.8 & 84.1 & 662 \\
\rowcolor[HTML]{ECF4FF} 
$\MODEL_{2}$ & 80.6 & 78.8 & 82.7 & 707 \\
claude-3-5-sonnet & 79.3 & 77.2 & 81.3 & 567 \\
gpt-4o-2024-05-13 & 79.2 & 77.3 & 80.9 & 696 \\
gpt-4-0125-preview & 78.0 & 75.9 & 80.4 & 619 \\
qwen2.5-72b-instruct & 78.0 & 76.2 & 79.8 & 821 \\
gpt-4o-2024-08-06 & 77.9 & 75.9 & 80.0 & 594 \\ \bottomrule
\end{tabular}
}
\caption{Results on full Arena-Hard dataset without style control.}
\label{exp:arena_wo}
\end{table}

\end{document}